# Global Weisfeiler-Lehman Kernels

Christopher Morris
TU Dortmund University
christopher.morris@tu-dortmund.de

Kristian Kersting
Technische Universität Darmstadt
kersting@cs.tu-darmstadt.de

Petra Mutzel
TU Dortmund University
petra.mutzel@tu-dortmund.de

*Abstract*—Most state-of-the-art graph kernels only take local graph properties into account, i.e., the kernel is computed with regard to properties of the neighborhood of vertices or other small substructures. On the other hand, kernels that do take global graph properties into account may not scale well to large graph databases. Here we propose to start exploring the space between local and global graph kernels, striking the balance between both worlds. Specifically, we introduce a novel graph kernel based on the $k$-dimensional Weisfeiler-Lehman algorithm. Unfortunately, the $k$-dimensional Weisfeiler-Lehman algorithm scales exponentially in $k$. Consequently, we devise a stochastic version of the kernel with provable approximation guarantees using conditional Rademacher averages. On bounded-degree graphs, it can even be computed in constant time. We support our theoretical results with experiments on several graph classification benchmarks, showing that our kernels often outperform the state-of-the-art in terms of classification accuracies.

## I. INTRODUCTION

In several domains like chemo- and bioinformatics, or social network analysis large amounts of structured data, i.e., *graphs*, are prevalent. Surprisingly, most state-of-the-art *graph kernels* only take *local* graph properties into account, i.e., they are computed based on properties of the neighborhood of vertices or other small substructures, e.g., see [1]–[4]. Moreover, kernels that do take global graph properties into account may not scale to large graph databases, e.g., see [5].

Recently, a graph kernel based on the 1-*dimensional Weisfeiler-Lehman* algorithm [6] has been proposed [3]. The 1-dimensional Weisfeiler-Lehman or *color refinement* algorithm is a well known heuristic for deciding whether two graphs are isomorphic: Given an initial *coloring* or *labeling* of the vertices of both graphs, e.g., their degree, in each iteration two vertices with the same label are assigned different labels if the number of identically labeled neighbors is not equal. If, after some iteration, the number of vertices with a certain label is different in both graphs, the algorithm terminates and we conclude that the two graphs are not isomorphic. It is easy to see that the algorithm is not able to distinguish all non-isomorphic graphs, e.g., see [7]. On the other hand, this simple algorithm is already quite powerful since it can distinguish almost all graphs [6], [8], and has been applied in other areas [9]–[13]. Moreover, it can be generalized to $k$-tuples leading to a more powerful graph isomorphism heuristic, which has been investigated in depth by the theoretical computer science community, e.g., see [7], [14], [15]. The so called *Weisfeiler-Lehman subtree kernel* is defined by computing the color refinement algorithm for a predefined number of steps and the kernel finally counts the number of common labels arising in all refinement steps.

Intuitively, the Weisfeiler-Lehman subtree kernel takes only local graph properties into account when performed for a fixed number of iterations. On the other hand, the $k$-dimensional variant does take more global properties into account but does not consider local properties.

### A. Contributions

Our contributions can be summed up as follows.

*a) A Kernel Based on the $k$-dimensional Weisfeiler-Lehman Algorithm:* We propose a graph kernel based on the $k$-dimensional Weisfeiler-Lehman algorithm. In order to take local properties into account we propose a *local* variant of the above algorithm, which can take local and global graph properties into account.

*b) An Approximation Algorithm for the Local $k$-dimensional Weisfeiler-Lehman Kernel:* Since the running time of the $k$-dimensional Weisfeiler-Lehman algorithm is in $\Omega(n^k)$ for fixed $k$, where $n$ denotes the number of nodes of the graph, our local kernel does not scale to large graph databases. Hence, we propose algorithms to approximate it. For bounded-degree graphs we show that the running time of the approximation algorithm is *constant*, i.e., it does not depend on the number of vertices or edges of a graph. Moreover, for general graphs, we propose an *adaptive* sampling algorithm which uses results from statistical learning theory, in particular, *conditional Rademacher averages*. Additionally, we present a procedure based on sparse linear algebra to compute our local kernel, which speeds up the computation time of the exact algorithms.

*c) Experimental Evaluation:* Finally, we implemented our algorithms and evaluated them on graph databases stemming from chemoinformatics and social network analysis. The results state that using global *and* local properties indeed improves performance over purely local or purely global approaches. Moreover, our kernels perform well over all data sets which is not the case for the other (purely local or global) kernels.

### B. Related Work

In recent years, various graph kernels have been proposed, e.g., see [16], [17]. Gärtner, Flach, and Wrobel [18] and Kashima, Tsuda, and Inokuchi [19] simultaneously proposed graph kernels based on random walks, which count the number of walks two graphs have in common. Since then, random walk kernels have been studied intensively [16], [20]–[23]. Kernels based on tree patterns were initially proposed by Ramon and Gärtner [24] and later refined by Mahé and Vert [25]. Kernels based on shortest paths were first proposed by Borgwardt and Kriegel [26], and are computed by performing one-step walks on transformed input graphs, where edges are annotated with shortest-path distances. A drawback of the approaches mentioned above is their *high computational cost*. They all employ the kernel trick, leading to a quadratic overhead in the size of the data set, and the running time for a single kernel function evaluation can be only bounded by a polynomial function of relatively high degree, e.g., the random-walk and the shortest-path graph kernel have a running time in $\mathcal{O}(n^{2\omega})$ and $\mathcal{O}(n^4)$, respectively, where $n$ denotes the maximum number of vertices of two graphs, and $\omega$ denotes the exponent of the running time of matrix multiplication. Moreover, recently graph kernels using matchings [27] and geometric embeddings [28], [29] have been proposed.

A different line in the development of graph kernels focused particularly on scalable graph kernels. These kernels are typically computed efficiently by explicit feature maps, which allow to bypass the computation of a gram matrix, and allow applying scalable linear classification algorithms [30]–[32]. Prominent examples are kernels based on subgraphs up to a fixed size, so called *graphlets* [1], [33], or specific subgraphs like cycles and trees [34]. Other approaches of this category encode the neighborhood of every node by different techniques, e.g., see [17], [35], [36], and most notably the Weisfeiler-Lehman subtree kernel [3]. Subgraph and Weisfeiler-Lehman kernels have been successfully employed within frameworks for smoothed and deep graph kernels [37], [38]. Moreover, procedures to speed-up the computation time of the Weisfeiler-Lehman subtree kernel in practice [10], a streaming version [39], a variant for dynamic graphs [40], and a variant that can handle continuous node and edge labels [41] have been proposed.

In the past, few works considered graph kernels that use global graph properties. In [5] a kernel based on the Lovàsz number is proposed. Moreover, Kondor and Pan [42] proposed a graph kernel based on the graph Laplacian.

Moreover, few works considered sampling as way to speed up graph kernel computation. In [1] a sampling algorithm for the graphlet kernel is introduced, which relies on approximating the relative counts of graphlets of a certain size. The authors provide a bound on the number of samples needed to approximate this quantity for a specific graphlet. However, they do not show an approximation result for the kernel. One drawback here is that the algorithm has to solve the graph isomorphism problem as a subproblem. More refined approaches can be found, e.g., in [43]–[45].

In [5] sampling techniques for approximating the kernel based on the Lovàsz number were used. However, the running time of the algorithm is at least quadratic in the number of vertices for a single evaluation of the kernel function.

## C. Outline

In Section II we fix some notation and describe the 1-dimensional Weisfeiler-Lehman algorithm and the corresponding kernel. Moreover, we describe a variant of the $k$-dimensional Weisfeiler-Lehman algorithm. In Section III section we propose our local kernel based on the above variant of the $k$-dimensional Weisfeiler-Lehman algorithm and propose a procedure to speed up the computation time of our kernels based on sparse matrix computation. Subsequently, we present our approximation algorithm for bounded-degree graphs and the approximation algorithm for general graphs based on conditional Rademacher averages. In Section V, we report on the results of our experimental evaluation.

## II. PRELIMINARIES

A *graph* $G$ is a pair $(V, E)$ with a *finite* set of *nodes* $V$ and a set of *edges* $E \subseteq \{\{u, v\} \subseteq V \mid u \neq v\}$. We denote the set of nodes and the set of edges of $G$ by $V(G)$ and $E(G)$, respectively. For ease of notation we denote the edge $\{u, v\}$ in $E(G)$ by $(u, v)$ or $(v, u)$. In the case of *directed graphs* $E \subseteq \{(u, v) \in V \times V \mid u \neq v\}$.

A *labeled graph* is a graph $G$ endowed with a *label function* $l \colon V(G) \to \Sigma$, where $\Sigma$ is some finite alphabet. We say that $l(v)$ is a *label* of $v$ for $v$ in $V(G)$.

Moreover, $N(v)$ denotes the *neighborhood* of $v$ in $V(G)$, i.e., $N(v) = \{u \in V(G) \mid (v, u) \in E(G)\}$. A graph is of *$d$-bounded degree* if its maximum degree is at most $d$, where $d$ is always independent of the number of vertices, i.e., $d$ is in $\mathcal{O}(1)$. We say that two graphs $G$ and $H$ are *isomorphic* if there exists an edge preserving bijection $\varphi \colon V(G) \to V(H)$, i.e., $(u, v)$ is in $E(G)$ if and only if $(\varphi(u), \varphi(v))$ is in $E(H)$. Let $S \subseteq V(G)$ then $G[S] = (S, E_S)$ is the *subgraph induced* by $S$ with $E_S = \{(u, v) \in E(G) \mid u, v \in S\}$.

Let $\chi$ be a non-empty set and let $k \colon \chi \times \chi \to \mathbb{R}$ be a function. Then $k$ is a *kernel* on $\chi$ if there is a real Hilbert space $\mathcal{H}_k$ and a feature map $\phi \colon \chi \to \mathcal{H}_k$ such that $k(x, y) = \langle \phi(x), \phi(y) \rangle$ for $x$ and $y$ in $\chi$, where $\langle \cdot, \cdot \rangle$ denotes the inner product of $\mathcal{H}_k$. Moreover, $\mathbf{K}$ in $\mathbb{R}^{n \times n}$ denotes the *gram matrix* for a kernel $k$ for some finite subset $S$ of $\chi$ of cardinality $n$, i.e., $\mathbf{K}_{ij} = k(x_i, x_j)$ for $x_i$ and $x_j$ in $S$. Let $\mathbb{G}$ be a non-empty set of graphs, then a kernel $k \colon \mathbb{G} \times \mathbb{G} \to \mathbb{R}$ is called *graph kernel*. Moreover, let $[1 \colon n] = \{1, \ldots, n\} \subset \mathbb{N}$ for $n > 1$, let $S$ be a set then the set of $k$-sets $S_k = \{U \subseteq S \mid |U| = k\}$ for $k \geq 2$, which is the set of all subsets with cardinality $k$, and let $\{\!\{ \ldots \}\!\}$ denote a multiset.

### A. Weisfeiler-Lehman Subtree Kernel

We now describe the 1-dimensional Weisfeiler-Lehman algorithm (1-WL) and the corresponding kernel. Let $G$ and $H$ be graphs, and let $l$ be a label function $V(G) \cup V(H) \to \Sigma$, e.g., $l(v) = |N(v)|$ for $v$ in $V(G) \cup V(H)$. In each iteration $i \geq 0$, the 1-WL algorithm computes a new label function $l^i \colon V(G) \cup V(H) \to \Sigma$. In iteration 0 we set $l^0 = l$. Now in iteration $i > 0$, we set

$$l^i(v) = \mathsf{relabel}((l^{i-1}(v), \mathsf{sort}(\{\!\{l^{i-1}(u) \mid u \in N(v)\}\!\}))),$$

for $v$ in $V(G) \cup V(H)$, where $\mathsf{sort}(S)$ returns a (ascendantly) sorted tuple of the multiset $S$ and the bijection $\mathsf{relabel}(p)$ maps the pair $p$ to an unique value in $\Sigma$, which has not been used in previous iterations. Now if $G$ and $H$ have an unequal number of vertices labeled $\sigma$ in $\Sigma$, we conclude that the graphs are not isomorphic. Moreover, if the cardinality of the image of $l^{i-1}$ equals the cardinality of the image of $l^i$ the algorithm terminates. After at most $|V(G)| + |V(H)|$ iterations the algorithm terminates. It is easy to see that the algorithm is not able to distinguish all non-isomorphic graphs, e.g., see [7]. On the other hand, this simple algorithm is already quite powerful, since it can distinguish almost all graphs [8]. The running time of the algorithm is in $\mathcal{O}(m \log n)$ where $n = |V(G)|+|V(H)|$ and $m = |E(G)| + |E(H)|$, e.g., see [46].

The idea of the Weisfeiler-Lehman subtree graph kernel is to compute the above algorithm for $h \geq 0$ iterations, and after each iteration $i$ compute a feature vector $\phi^i(G)$ in $\mathbb{R}^{|\Sigma_i|}$ for each graph $G$, where $\Sigma_i \subseteq \Sigma$ denotes the image of $l^i$. Each component $\phi^i(G)_{\sigma_j^i}$ counts the number of occurrences of vertices labeled with $\sigma_j^i$ in $\Sigma_i$. The overall feature vector $\phi(G)$ is defined as the concatenation of the feature vectors of all $h$ iterations, i.e.,

$$\left(\phi^0(G)_{\sigma_1^0}, \ldots, \phi^0(G)_{\sigma_{|\Sigma_0|}^0}, \ldots, \phi^h(G)_{\sigma_1^h}, \ldots \phi^h(G)_{\sigma_{|\Sigma_h|}^h}\right).$$

Then the Weisfeiler-Lehman subtree kernel for $h$ iterations is $k_{\mathrm{WL}}(G,H) = \langle \phi(G), \phi(H) \rangle$. Moreover, let $k_{\mathrm{WL}}^h(G,H) = \langle \phi^h(G), \phi^h(H) \rangle$. The running time for a single feature vector computation is in $\mathcal{O}(hm)$ and $\mathcal{O}(Nhm + N^2hn)$ for the computation of the gram matrix for a set of $N$ graphs [3], where $n$ and $m$ denote the maximum number of vertices and edges over all $N$ graphs, respectively.

### B. The $k$-dimensional Weisfeiler-Lehman Algorithm

Based on the 1-WL, we consider the following variant[1] of the $k$-dimensional Weisfeiler-Lehman algorithm ($k$-WL) for $k \geq 2$. Let $G$ and $H$ be graphs.[2] Instead of iteratively labeling vertices, the $k$-WL computes a labeling function defined on the set of $k$-sets $V(G)_k \cup V(H)_k$. In order to describe the algorithm, we define the neighborhood $N(t)$ of a $k$-set $t = \{t_1, \ldots, t_k\}$ in $V(G)_k$ (analogously for $V(H)_k$):

$$N(t) = \{\{t_1, \ldots, t_{j-1}, r, t_{j+1}, \ldots, t_k\} \mid r \in V(G) \setminus \{t_j\}\}. \quad (1)$$

That is, the neighborhood $N(t)$ of $t$ is obtained by replacing a vertex $t_j$ from $t$ by a vertex from $V(G) \setminus \{t_j\}$, see Figure 1a for a graphical illustration. In iteration 0, the algorithm labels each $k$-set with its isomorphism type, i.e., two $k$-sets $s$ and $t$ in $V(G)_k$ get the same label if the corresponding induced subgraphs are isomorphic. Now in iteration $i > 0$ we set

$$l^i(t) = \mathsf{relabel}((l^{i-1}(t), \mathsf{sort}(\{\!\{l^{i-1}(s) \mid s \in N(t)\}\!\}))),$$

[1] In theoretical computer science it is usually defined on $k$-tuples instead of $k$-sets. Due to scalability we consider $k$-sets instead.

[2] For clarity of presentation we omit the presence of node and edge labels. Observe that the algorithm can be extended to take node and edge labels into account.

where $\mathsf{sort}(S)$ returns a (ascendantly) sorted tuple of the multiset $S$ of labels. The algorithm then proceeds analogously to the 1-WL.

## III. A LOCAL KERNEL BASED ON THE $k$-DIMENSIONAL WEISFEILER-LEHMAN ALGORITHM

The $k$-WL, see the Introduction and Section II-B, is inherently *global*, i.e., it labels a $k$-set $t$ by considering $k$-sets whose vertices are not connected to the vertices of $t$. Moreover, it does not take the sparsity of the underlying graph into account. In order to capture the local properties of a graph, we propose a *local* variant. The idea of our *local* $k$-WL ($k$-LWL) is the following: The algorithm again labels all $k$-sets. But in order to extract local features we define the *local* neighborhood of a $k$-set $t = \{t_1, \ldots, t_k\}$ in $V(G)_k$,

$$N^L(t) = \{\{t_1, \ldots, t_{j-1}, r, t_{j+1}, \ldots, t_k\} \mid r \in V(G) \setminus \{t_j\},$$
$$\text{and } \exists l \in [1\!:\!k]\colon (t_l, r) \in E(G)\},$$

i.e., we consider a $k$-set $s$ a local neighbor of a $k$-set $t$ if $s$ is in $N(t)$, and there is at least one edge between vertices of $s$ and $t$, see Figure 1b for a graphical illustration. Now the local algorithm works the same way as the $k$-WL but in each iteration $i > 0$ considers the local neighbors of a $k$-set, and computes a labeling function $l_\mathrm{L}^i \colon V(G)_k \to \Sigma$.

Moreover, in the following we define the notion of a (directed) $k$-set graph and the $c$-neighborhood of a $k$-set.

**Definition 1.** *Let $G$ be a graph, and let $s$ and $t$ be $k$-sets from $V(G)_k$, then the (directed) $k$-set graph $S(G) = (V_S, E_S)$, where $V_S = \{v_r \mid r \in V(G)_k\}$, and*

$$(v_s, v_t) \in E_S \iff t \in N^L(s).$$

**Definition 2.** *Let $G$ be a graph, and let $t$ be a $k$-set from $V(G)_k$, then the $c$-neighborhood of $t$ for $c \geq 0$,*

$$\mathbf{N}(t, c) = \{s \in V(G)_k \mid d(v_t, v_s) \leq c\},$$

*where $d \colon V_S \times V_S \to \mathbb{N}$ denotes the shortest-path distance in the $k$-set graph $S(G)$ of $G$.*

### A. A Kernel Based on the $k$-LWL

The idea for a kernel based on the $k$-LWL is the same as for the 1-WL. We compute the $k$-LWL for $h \geq 0$ iterations, and after each iteration $i$ compute a feature vector $\phi_{k\text{-LWL}}^i(G)$ in $\mathbb{R}^{|\Sigma_i|}$ for each graph $G$, where $\Sigma_i \subseteq \Sigma$ denotes the image of $l_\mathrm{L}^i$. Each component $\phi_{k\text{-LWL}}^i(G)_{\sigma_j^i}$ counts the number of occurrences of $k$-sets labeled with $\sigma_j^i$ in $\Sigma_i$. The overall feature vector $\phi_{k\text{-LWL}}(G)$ is defined as the concatenation of the feature vectors of all $h$ iterations, i.e.,

$$\left(\phi^0(G)_{\sigma_1^0}, \ldots, \phi^0(G)_{\sigma_{|\Sigma_0|}^0}, \ldots, \phi^h(G)_{\sigma_1^h}, \ldots \phi^h(G)_{\sigma_{|\Sigma_h|}^h}\right).$$

Let $G$ and $H$ be two graphs then the kernel $k_{k\text{-LWL}} = \langle \phi_{k\text{-LWL}}(G), \phi_{k\text{-LWL}}(H) \rangle$ and $k_{k\text{-LWL}}^h = \langle \phi_{k\text{-LWL}}^h(G), \phi_{k\text{-LWL}}^h(H) \rangle$.

Moreover, we will need the *normalized feature vector*

$$\widehat{\phi}_{k\text{-LWL}}^h(G) = \phi_{k\text{-LWL}}^h(G) / \|\phi_{k\text{-LWL}}^h(G)\|_1. \quad (2)$$

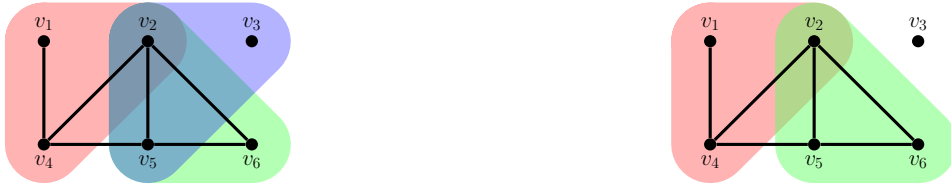

(a) Illustration of the global neighborhood of the $k$-set $\{v_2, v_4, v_5\}$, for better clarity we only depict a subset.

(b) Illustration of the local neighborhood of the $k$-set $\{v_2, v_4, v_5\}$, for better clarity we only depict a subset.

Fig. 1: Illustration of the local and global neighborhood of the $k$-set $\{v_2, v_4, v_5\}$

Observe that the components of the normalized feature vector are in $[0, 1]$. We denote the corresponding kernel by $\widehat{k}_{k\text{-LWL}}^h$.

*B. Sparse Linear Algebra Implementation*

Kersting, Mladenov, Garnett, and Grohe [10] proposed a linear algebra based implementation of the 1-WL: Let $G = (\{v_1, \ldots, v_n\}, E)$ be a graph with adjacency matrix $A$, let $c^0 = (l^0(v_1), \ldots, l^0(v_n))$. Moreover, let $S$ be a finite subset of the set of real numbers, and $\pi \colon S \to \mathbb{N}$ be a bijection that maps each element of $S$ to a prime number, then

$$c^h = \log(\pi(c^{h-1})) + A \log(\pi(c^{h-1})), \text{ and } l^h(v_i) = c_i^h$$

for $h > 0$, where we apply $\log$ and $\pi$ componentwise. This can be extended to the $k$-LWL by considering the adjacency matrix of the $k$-set graph. Observe that for the $k$-LWL a sparse graph will lead to a sparse adjacency matrix of the $k$-set graph, i.e., the sparsity only depends on the sparsity of the underlying graph and $k$.

## IV. APPROXIMATION ALGORITHMS FOR THE LOCAL $k$-DIMENSIONAL WEISFEILER-LEHMAN KERNEL

Since the worst-case running time of the computation of $\phi_{k\text{-LWL}}^h(G)$ can be lower bounded by $|V(G)_k|$ for fixed $k$, computing the corresponding kernel is not feasible for large graphs. Thereto, we describe approximation algorithms for the $k$-LWL kernel. First, we describe an algorithm restricted to bounded-degree graphs, and show that it can be computed in *constant time*. Subsequently, in Section IV-B, we describe an algorithm for general graphs which employs an adaptive sampling strategy.

*A. A Sampling Algorithm for Bounded-degree Graphs*

The idea of the first algorithm is the following: Let $G$ be a graph and let $\widetilde{\phi}_{k\text{-LWL}}^h(G)$ be a zero vector with the same number of components as $\widehat{\phi}_{k\text{-LWL}}^h(G)$. First, we sample a multiset $S$ of $k$-sets, where each element is sampled independently and uniformly at random from $V(G)_k$. The exact cardinality of $S$ will be determined later. Secondly, for each such $k$-set $s$ in $S$, we compute the $h$-neighborhood $\mathbf{N}(s, h)$ and compute the $k$-LWL on the subgraph induced by all vertices in

$$\mathfrak{N}(s, h) = \{u, v, \ldots, w \mid \{u, v, \ldots, w\} \in \mathbf{N}(s, h)\}$$

**Algorithm 1** Approximation Algorithm for the $k$-LWL for Bounded-degree Graphs

**Require:** A $d$-bounded degree graph $G$, number of iterations $h \geq 0$, $k \geq 2$, failure parameter $\delta$ in $(0, 1)$, and an additive error term $\varepsilon$ in $(0, 1]$.
**Ensure:** A feature map $\widetilde{\phi}_{k\text{-LWL}}^h(G)$ according to Inequality (4).
1: Let $\widetilde{\phi}_{k\text{-LWL}}^h(G)$ be a feature vector
2: Draw a multiset $S$ of $k$-sets independently and uniformly from $V(G)_k$ with $|S|$ according to Equation (3).
3: **parallel for** $s \in S$ **do**
4:     Compute the $h$-neighborhood $\mathbf{N}(s, h)$ around $s$
5:     Compute $\sigma = l_{L,\mathfrak{N}}^h(s)$ on $G[\mathfrak{N}(s, h)]$
6:     $\widetilde{\phi}_{k\text{-LWL}}^h(G)_\sigma = \widetilde{\phi}_{k\text{-LWL}}^h(G)_\sigma + 1/|S|$
7: **end**
8: **return** $\widetilde{\phi}_{k\text{-LWL}}^h(G)$

resulting in a label $\sigma$ for the $k$-set $s$. The following result shows that the label of a $k$-set after $h$ iterations can be computed *locally* by only considering its $h$-neighborhood.

**Lemma 1.** *Let $G$ be a graph, and let $s$ be a $k$-set in $V(G)_k$. Moreover, let $l_{L,\mathfrak{N}}^h(s)$ be the label of $s$ after the $h$-th iteration of the $k$-LWL on $G[\mathfrak{N}(s, h)]$, then $l_{L,\mathfrak{N}}^h(s) = l_L^h(s)$.*

*Proof (Sketch).* Induction on the number of iterations. □

Now the algorithm proceeds by adding $1/|S|$ to $\widetilde{\phi}_{k\text{-LWL}}^h(G)_\sigma$. An easy implication of the above lemma is that for $k$-sets that are in $\mathbf{N}(t, i)$ for $i \leq h$ it holds that $l_{L,\mathfrak{N}}^j(t) = l_L^j(t)$ for $j \leq h$ such that $i + j \leq h$. See Algorithm 1 for pseudo code. Note that lines 4 to 6 can be computed in parallel for all samples. Moreover, note that the algorithm can be easily adapted so that it approximates the feature vector over all $h$ iterations. Let $G$ be a $d$-bounded degree graph, and let $\Gamma(d, h)$ be an upper bound on the maximum number of different labels of the $k$-LWL after $h$ iterations on $G$. We get the following result.

**Theorem 1.** *Let $G$ be a $d$-bounded degree graph, and let*

$$|S| \geq \left\lceil \frac{\log(2 \cdot \Gamma(d, h) \cdot 1/\delta)}{2(\varepsilon/\Gamma(d, h))^2} \right\rceil. \quad (3)$$

Then Algorithm 1 approximates the normalized feature vector $\widehat{\phi}_{k\text{-LWL}}^{h}(G)$ of the $k$-LWL such that with probability $(1-\delta)$ for $\delta$ in $(0,1)$,

$$\left\|\widehat{\phi}_{k\text{-LWL}}^{h}(G) - \widetilde{\phi}_{k\text{-LWL}}^{h}(G)\right\|_{1} \leq \varepsilon \quad (4)$$

for any $\varepsilon$ in $(0,1]$. Morever, the running time of the algorithm is only dependent of $d$, $k$, and $h$, i.e., it does not depend on $|V(G)|$.

*Proof.* First, observe that $\Gamma(d,h)$ is only dependent on the number of iterations $h$, $k$, and the maximum degree $d$.

Let $X_{i,\sigma}$ denote the random variable that is 1 if we sample a $k$-set $s$ such that $l_{L,\mathfrak{N}}^{h}(s) = \sigma$ in iteration $i$ of Algorithm 1, otherwise 0. Now observe that

$$\mathbb{E}(X_{i,\sigma}) = \widehat{\phi}_{k\text{-LWL}}^{h}(G)_{\sigma}.$$

Moreover, let $\bar{X}_{\sigma} = 1/|S| \cdot \sum_{i=1}^{S} X_{i,\sigma} = \widetilde{\phi}_{k\text{-LWL}}^{h}(G)_{\sigma}$, then, by the linearity of expectation,

$$\mathbb{E}(\bar{X}_{\sigma}) = \widehat{\phi}_{k\text{-LWL}}^{h}(G)_{\sigma}.$$

Hence, by the Hoeffding bound [47], we get

$$\mathbb{P}\Big(\big|\bar{X}_{\sigma} - \widehat{\phi}_{k\text{-LWL}}^{h}(G)_{\sigma}\big| \geq \lambda\Big) \leq 2e^{-2|S|\cdot\lambda^{2}}.$$

By setting the sample size

$$|S| \geq \left\lceil \frac{\log(2 \cdot \Gamma(d,h) \cdot 1/\delta)}{2\lambda^{2}} \right\rceil, \quad (5)$$

it follows that

$$\mathbb{P}\Big(\big|\bar{X}_{\sigma} - \widehat{\phi}_{k\text{-LWL}}^{h}(G)_{\sigma}\big| \geq \lambda\Big) \leq \frac{\delta}{\Gamma(d,h)}.$$

The result then follows by setting $\lambda = \varepsilon/\Gamma(d,h)$, and the Union bound. Finally the bound on the running time follows from the observation that Equation (3), and the running time of lines 4 to 6 in Algorithm 1 are independent of the size of $G$, i.e., the number of vertices and edges. The correctness follows from Lemma 1. $\square$

Now let $\widetilde{k}_{k\text{-LWL}}^{h}(G,H)$ denote the corresponding kernel, i.e., $\widetilde{k}_{k\text{-LWL}}^{h}(G,H) = \langle \widetilde{\phi}_{k\text{-LWL}}^{h}(G), \widetilde{\phi}_{k\text{-LWL}}^{h}(H)\rangle$ for two graphs $G$ and $H$. The following proposition shows that the above kernel approximates the normalized $k$-LWL kernel arbitrarily close.

**Proposition 1.** *Let $\mathbb{G}$ be (non-empty, finite) set of $d$-bounded degree graphs, let $\widehat{k}_{k\text{-LWL}}^{h}$ be the normalized $k$-LWL kernel, and let*

$$|S| \geq \left\lceil \frac{\log(2 \cdot \Gamma(d,h) \cdot 1/\delta \cdot |\mathbb{G}|)}{2(\lambda/\Gamma(d,h))^{2}} \right\rceil.$$

*Then with probability $(1-\delta)$ for $\delta$ in $(0,1)$, Algorithm 1 approximates $\widehat{k}_{k\text{-LWL}}^{h}$ such that*

$$\sup_{G,H \in \mathbb{G}} \left|\widehat{k}_{k\text{-LWL}}^{h}(G,H) - \widetilde{k}_{k\text{-LWL}}^{h}(G,H)\right| \leq 3\lambda$$

*for any $\lambda$ in $(0,1]$. The running time for computing $\widetilde{\mathbf{K}}_{k\text{-LWL}}^{h}$ for $\mathbb{G}$ does only depend on the cardinality of $\mathbb{G}$, the number of iterations $h$, $k$, and the maximum degree $d$.*

*Proof.* First observe that by setting the sample size $|S|$ to

$$|S| \geq \left\lceil \frac{\log(2 \cdot \Gamma(d,h) \cdot 1/\delta \cdot |\mathbb{G}|)}{2\varepsilon^{2}} \right\rceil,$$

we get that with probability $(1-\delta)$ for all $G$ in $\mathbb{G}$

$$\left|\widehat{\phi}_{k\text{-LWL}}^{h}(G)_{i} - \widetilde{\phi}_{k\text{-LWL}}^{h}(G)_{i}\right| \leq \varepsilon$$

for any $1 \leq i \leq \Gamma(d,h)$ holds. Let $G$ and $H$ in $\mathbb{G}$, then

$$\widetilde{k}_{k\text{-LWL}}^{h}(G,H) = \left\langle \widetilde{\phi}_{k\text{-LWL}}^{h}(G), \widetilde{\phi}_{k\text{-LWL}}^{h}(H) \right\rangle$$

$$= \sum_{i=1}^{\Gamma(d,h)} \widetilde{\phi}_{k\text{-LWL}}^{h}(G)_{i} \cdot \widetilde{\phi}_{k\text{-LWL}}^{h}(H)_{i}$$

$$\leq \sum_{i=1}^{\Gamma(d,h)} \left(\widehat{\phi}_{k\text{-LWL}}^{h}(G)_{i} + \varepsilon\right) \cdot \left(\widehat{\phi}_{k\text{-LWL}}^{h}(H)_{i} + \varepsilon\right)$$

$$\leq \sum_{i=1}^{\Gamma(d,h)} \left(\widehat{\phi}(G)_{i} \cdot \widehat{\phi}(H)_{i}\right) + \varepsilon \cdot \sum_{i=1}^{\Gamma(d,h)} \left(\widehat{\phi}(G)_{i} + \widehat{\phi}(H)_{i}\right)$$

$$+ \sum_{i=1}^{\Gamma(d,h)} \varepsilon^{2} \leq \widehat{k}_{k\text{-LWL}}^{h}(G,H) + 2\Gamma(d,h) \cdot \varepsilon + \Gamma(d,h) \cdot \varepsilon.$$

The last inequality follows from the fact that the components of $\widehat{\phi}(\cdot)$ are in $[0,1]$. The result then follows by setting $\varepsilon = \lambda/\Gamma(d,h)$. $\square$

Note that the above technique also leads to an approximation result for the (normalized) Weisfeiler-Lehman subtree kernel.

**Corollary 1.** *Let $\mathbb{G}$ be (non-empty, finite) set of $d$-bounded degree graphs, let $\widehat{k}_{WL}^{h}$ be the normalized Weisfeiler-Lehman subtree kernel, and let*

$$|S| \geq \left\lceil \frac{\log(2 \cdot \Gamma(d,h) \cdot 1/\delta \cdot |\mathbb{G}|)}{2(\lambda/\Gamma(d,h))^{2}} \right\rceil.$$

*Then with probability $(1-\delta)$ for $\delta$ in $(0,1)$, Algorithm 1 approximates $\widehat{k}_{WL}^{h}$ such that*

$$\sup_{G,H \in \mathbb{G}} \left|\widehat{k}_{WL}^{h}(G,H) - \widetilde{k}_{WL}^{h}(G,H)\right| \leq 3\lambda$$

*for any $\lambda$ in $(0,1]$. The running time for computing $\widetilde{\mathbf{K}}_{WL}^{h}$ for $\mathbb{G}$ does only depend on the size of $\mathbb{G}$, the number of iterations $h$, and the maximum degree $d$.*

Observe that instead of considering bounded-degree graphs Proposition 1 and Corollary 1 also hold for general graphs with a constant label alphabet, i.e., the cardinality of the label alphabet is independent of the size of the graphs. Moreover, observe that Proposition 1 and Corollary 1 can be easily adapted such that we get approximation results for the kernels over all $h$ iterations.

## B. Adaptive Sampling Algorithm for General Graphs

In order to calculate the sample size of Algorithm 1, we assumed to know the maximum degree or the number of different labels in advance. Hence, in order to circumvent this problem, we propose a variant of Algorithm 1 that relies on an adaptive sampling scheme based on *conditional Rademacher averages*.

Let $\mathcal{D}$ be a finite set, and let $\mathcal{F} = \{f \mid \mathcal{D} \to [0,1]\}$ be a family of functions. Now let $S = \{s_1, \ldots, s_m\}$ be a sample of elements sampled independently and uniformly from $\mathcal{D}$. Next we define the *true average* and the *sample average*, respectively,

$$L_\mathcal{D}(f) = \mathbb{E}_{s \sim \mathcal{U}(\mathcal{D})}[f(s)], \quad \text{and} \quad L_S(f) = \frac{1}{m}\sum_{i=1}^m f(s_i).$$

Given $S$ we are interested in bounding

$$\sup_{f \in \mathcal{F}} |L_\mathcal{D}(f) - L_S(f)|, \qquad (6)$$

i.e., determining the maximum deviation of the sample average to the true average using only the sample $S$. For $i$ in $[1\!:\!m]$ let $\sigma_i$ be a *Rademacher variable*, i.e., it is i.i.d. to $\mathbb{P}[\sigma_i = 1] = \mathbb{P}[\sigma_i = -1] = 1/2$. The conditional Rademacher average [48] is defined as

$$\mathcal{R}_\mathcal{F}(S) = \frac{1}{m}\mathbb{E}_{\sigma \sim \{\pm 1\}^m}\left[\sup_{f \in \mathcal{F}} \sum_{i=1}^m \sigma_i f(s_i)\right]. \qquad (7)$$

We can now state a classical result from learning theory which employs Equation (7) to bound Equation (6) depending on the sample $S$.

**Theorem 2** ([48]). *Let $\delta$ be in $(0,1)$, then with probability of at least $(1-\delta)$,*

$$\sup_{f \in \mathcal{F}} |L_\mathcal{D}(f) - L_S(f)| \leq 2 \cdot \mathcal{R}_\mathcal{F}(S) + 3\sqrt{\frac{\log(2/\delta)}{2m}}. \qquad (8)$$

This bound can be further improved [49]. In order to bound the conditional Rademacher average we can use *Massart's Lemma*. Let

$$\mathbf{v}_{f,S} = (f(s_1), \ldots, f(s_m))$$

for $f$ in $\mathcal{F}$, and let the set $\mathcal{V}_S = \{\mathbf{v}_{f,S} \mid f \in \mathcal{F}\}$. We get the following result.

**Lemma 2** ([48]).

$$\mathcal{R}_\mathcal{F}(S) \leq \max_{f \in \mathcal{F}} \|\mathbf{v}_{f,S}\| \frac{\sqrt{2\ln|\mathcal{V}_S|}}{m}.$$

Moreover, based on the above lemma Riondato and Upfal [50] proposed a convex optimization problem to achieve tighter (empirical) bounds on the conditional Rademacher average.

In order to use the above theorem to approximate the feature vector of the kernel based on the $k$-LWL, we define the following family of functions

$$\mathcal{F}_{G,k\text{-LWL}} = \{\mathbf{1}_{\sigma,h} \mid \sigma \in \Sigma_h, \text{ and } h \geq 0\}$$

for a graph $G$, where $\mathbf{1}_{\sigma,h}$ equals 1 for a $k$-set $t$ in $V(G)_k$ if it has label $\sigma$ after $h$ iterations of the $k$-LWL, and otherwise 0, i.e.,

$$\mathbf{1}_{\sigma,h}(t) = \begin{cases} 1 & \text{if } l_\text{L}^h(t) = \sigma, \\ 0 & \text{otherwise}. \end{cases}$$

Observe that

$$L_\mathcal{D}(\mathbf{1}_{\sigma,h}) = \widehat{\phi}(G)_\sigma^h.$$

Moreover, observe that $\mathbf{v}_{S,f}$ for $f$ in $\mathcal{F}_{G,k\text{-LWL}}$ and $|\mathcal{V}_S|$ can be computed from $S$. Notice that $|\mathcal{V}_S|$ will be usually much smaller than $|\mathcal{F}_{G,k\text{-LWL}}|$. The idea of the sampling algorithm is the following: In each iteration $i \geq 0$ we determine a sample size $|S_i|$. Draw $|S_i|$ samples independently and uniformly at random from $V(G)_k$, and proceed as in Algorithm 1, i.e., we compute the $h$-neighborhood of each $k$-set $t$ in $S_i$, and compute its label $l_{\text{L},\mathfrak{N}}^h(t)$ after $h$ iterations on the induced subgraph of the set $\mathfrak{N}(t,h)$. The algorithm terminates if the right-hand side of Inequality (8) together with the bound of Lemma 2 is less or equal than $\varepsilon$ in $(0,1]$, otherwise we proceed with iteration $i+1$. See Algorithm 2 for pseudo code.

---

**Algorithm 2** Adaptive Approximation Algorithm for the $k$-LWL

**Require:** A graph $G$, number of iterations $h \geq 0$, $k \geq 2$, failure parameter $\delta$ in $(0,1)$, and an additive error term $\varepsilon$ in $(0,1]$.
**Ensure:** A feature map $\widetilde{\phi}_{k\text{-LWL}}^h(G)$ according to Theorem 3.
1: Let $\widetilde{\phi}_{k\text{-LWL}}^h(G)$ be a feature vector
2: $i \leftarrow 0$, $s \leftarrow 0$

3: **while** Inequality (8) not satisfied **do**
4:     Compute $S_i$
5:     $s \leftarrow s + |S_i|$
6:     **parallel for** $t \in S_i$ **do**
7:         Compute the $h$-neighborhood $\mathbf{N}(t,h)$ around $t$
8:         Compute $l_{\text{L},\mathfrak{N}}^h(t) = \sigma$ on $G[\mathfrak{N}(t,h)]$
9:         $\widetilde{\phi}_{k\text{-LWL}}^h(G)_\sigma = \widetilde{\phi}_{k\text{-LWL}}^h(G)_\sigma + 1$
10:     **end**
11:     $i \leftarrow i + 1$
12: **end while**

13: $\widetilde{\phi}_{k\text{-LWL}}^h(G) = 1/s \cdot \widetilde{\phi}_{k\text{-LWL}}^h(G)$
14: **return** $\widetilde{\phi}_{k\text{-LWL}}^h(G)$

---

We get the following result.

**Theorem 3.** *Let $G$ be a graph, then Algorithm 2 approximates the normalized feature vector $\widehat{\phi}_{k\text{-LWL}}^h(G)$ of the $k$-LWL such that with probability $(1-\delta)$ for $\delta$ in $(0,1)$,*

$$\sup_{\sigma \in \Sigma_h} \left| \widehat{\phi}_{k\text{-LWL}}^h(G)_\sigma - \widetilde{\phi}_{k\text{-LWL}}^h(G)_\sigma \right| \leq \varepsilon, \qquad (9)$$

*for any $\varepsilon$ in $(0,1]$.*

*Proof.* The finiteness follows from the fact that we increase the sample size in every iteration. Hence, eventually Inequality (8) will hold. The approximation guarantee then follows from Algorithm 2, Inequality (8), and the correctness of Algorithm 1. □

The above algorithm and the above result can be adapted straightforwardly to a set of graphs instead of a single graph.

## V. Experimental Evaluation

Our intention here is to investigate the benefits of the $k$-LWL kernel and the adaptive sampling algorithm compared to the state-of-the-art. More precisely, we address the following questions:

**Q1** How does the $k$-LWL kernel compare to state-of-the-art graph kernels in terms of classification accuracy?

**Q2** How does the $k$-LWL kernel compare to the (global) $k$-WL in terms of classification accuracy and running time?

**Q3** Does our adaptive sampling algorithm to approximate the $k$-LWL kernel speed up the computation time of the kernel computation?

**Q4** Does the adaptive sampling algorithm lead to worse classification accuracies compared to the exact algorithm?

**Q5** Does the linear algebra based implementation of the $k$-LWL kernel lead to a speed up in computation time? Does parallelization speed up the computation time?

### A. Data Sets and Graph Kernels

We used the following data sets to evaluate our kernels, see Table I for properties and statistics.[3]

**ENZYMES and PROTEINS** contain graphs representing proteins according to the graph model of [51]. Each node is annotated with a discrete label. The data sets are subdivided into six and two classes, respectively. Note that this is the same data set as used in [52], which does not contain all the annotations described and used in [51].

**IMDB-BINARY** is a movie collaboration data set first used in [38] based on data from IMDB[4]. Each node represents an actor or an actress, and there exists an edge between two nodes if the corresponding actor or actress appears in the same movie. The nodes are unlabeled. Each graph represents an ego network of an actor or actress. The nodes are unlabeled and the data set is divided into two classes corresponding to action or romantic movies.

**MUTAG** is a data set consisting of mutagenetic aromatic and heteroaromatic nitro compounds [53], [54] with seven discrete node labels.

**NCI1 and NCI109** are (balanced) subsets of data sets made available by the National Cancer Institute[5] [3], [33], consisting of chemical compounds screened for activity against non-small cell lung cancer and ovarian cancer cell lines, respectively. The nodes are annotated with discrete labels.

**PTC_FM** is a data set from the Predicte Toxicology Challenge (PTC)[6] containing chemical compounds labeled according to carcinogenicity on female mice (FM). The nodes are annotated with discrete labels. It is divided into two classes.

**REDDIT-BINARY** is a social network data set based on data from the content-aggregation website Reddit[7] [38]. Each node represents a user and two nodes are connected by an edge if one user responded to the other users comment. The nodes are unlabeled and the data set is divided into two classes representing question-answer-based or discussion-based communities.

We implemented the $\{2, 3\}$-LWL kernel and the adaptive sampling algorithm for the 3-LWL kernel (3-LWL-APPROX). Moreover, for the $\{2, 3\}$-LWL kernel we implemented the linear algebra based algorithm ($\{2, 3\}$-LWL-LA) described in Section III-B. Moreover, we implemented a parallel version of the 3-LWL kernel, where we parallelized the label computation in each iteration (3-LWL-PAR) over all $k$-sets, and a parallel version of the 3-LWL-APPROX kernel (3-LWL-APPROX-PAR), where we computed the colors of the sampled $k$-sets in parallel.

We compare our kernels to the Weisfeiler-Lehman subtree kernel [3], the graphlet kernel [1], and the shortest-path kernel [26]. Moreover, we implemented a kernel ($k$-GWL) based on the (global) $k$-WL for $k$ in $\{2, 3\}$ that uses the definition of neighborhood of Equation (1). Here we also implemented the linear algebra based algorithm ($\{2, 3\}$-GWL-LA). All kernels were (re-)implemented in C++11.[8]

### B. Experimental Protocol

For each kernel, we computed the normalized gram matrix. We computed the classification accuracies using the $C$-SVM implementation of LIBSVM [55], using 10-fold cross validation. The $C$-parameter was selected from $\{10^{-3}, 10^{-2}, \ldots, 10^2, 10^3\}$ by 10-fold cross validation on the training folds. For the 3-LWL-APPROX we set $\varepsilon$ of Inequality (9) to 0.05 (0.1), and the initial sample size was set to 100. We doubled the sampled size when Inequality (8) was not fulfilled.

We repeated each 10-fold cross validation ten times with different random folds, and report average accuracies and standard deviations. Since the adaptive approximation algorithm is a randomized algorithm, we computed each gram matrix ten times and report average classification accuracies, standard deviations, and running times. We report running times for the 1-WL, the $\{2, 3\}$-GWL, the $\{2, 3\}$-LWL, and the 3-LWL-APPROX, and the corresponding linear algebra based and parallel variants with five refinement steps. For the graphlet kernel we counted (labeled) connected subgraphs of size three. For measuring the classification accuracy the number of iterations of the 1-WL, $\{2, 3\}$-GWL, $\{2, 3\}$-LWL, and 3-LWL-APPROX were selected from $\{0, \ldots, 5\}$ using 10-fold cross validation on the training folds only.

---

[3] All data sets can obtained from http://graphkernels.cs.tu-dortmund.de.
[4] https://www.imdb.com/
[5] https://www.cancer.gov/
[6] https://www.predictive-toxicology.org/ptc/
[7] https://www.reddit.com/
[8] The source code can be obtained from https://github.com/chrsmrrs/glocalwl.

TABLE I: Data set statistics and properties.

| Data Set | Properties | | | | |
|---|---|---|---|---|---|
| | Number of Graphs | Number of Classes | ⌀ Number of Nodes | ⌀ Number of Edges | Node Labels |
| ENZYMES | 600 | 6 | 32.6 | 62.1 | ✓ |
| IMDB-BINARY | 1000 | 2 | 19.8 | 96.5 | ✗ |
| MUTAG | 188 | 2 | 17.9 | 19.8 | ✓ |
| NCI1 | 4110 | 2 | 29.9 | 32.3 | ✓ |
| NCI109 | 4127 | 2 | 29.7 | 32.1 | ✓ |
| PTC_FM | 349 | 2 | 14.1 | 14.5 | ✓ |
| PROTEINS | 1113 | 2 | 39.1 | 72.8 | ✓ |
| REDDIT-BINARY | 2000 | 2 | 429.6 | 497.8 | ✗ |

All experiments were conducted on a workstation with an Intel Core i7-3770 with 3.40GHz and 16GB of RAM running Ubuntu 16.04.5 LTS. The parallel algorithms were executed on four cores using `std::thread`. Moreover, we used GNU C++ Compiler 5.4.1 with the flag –O2.

*C. Results and Discussion*

In the following we answer questions **Q1** to **Q5**. See also Tables II and III.

**A1** On five out of eight data sets the 3-LWL or the 3-LWL-APPROX performs as good as or better than the state-of-the-art, e.g., on the challenging ENZYMES data set the 3-LWL performs more than 8% better than the 1-WL. Moreover, on the other data sets the 3-LWL is at most 4.5% worse than the best performing kernel. Observe that the 3-LWL performs well over all data sets.

**A2** On all data sets besides PTC_FM and MUTAG the 3-LWL achieves better classification accuracies than the 3-GWL. This is further illustrated by the much lower average ranking of the $k$-LWL compared to the $k$-GWL kernels. On the ENZYMES data set the 3-LWL is more than 12% better. This shows the benefit of using local as well as global features. On all data sets the 3-LWL is faster, e.g., on the ENZYMES data set, the 3-LWL is more than five times faster. On the PROTEINS and the REDDIT-BINARY data set the 3-GWL did not finish within one day or was out of memory. On these data sets the 3-LWL-APPROX was able to finish within one day. The same is true for the 2-LWL compared to the 2-GWL with regard to the REDDIT-BINARY data set.

**A3** The approximation algorithm speeds up the computation time of the kernel on the larger data sets (PROTEINS and REDDIT-BINARY). For example on the PROTEINS data set (with inital sample size 100 and $\varepsilon = 0.1$) it is over 126 times faster than the exact algorithm.

**A4** On all data sets excluding ENZYMES the 3-LWL-APPROX is at most 5% (with inital sample size 100 and $\varepsilon = 0.05$) worse than the exact algorithm in terms of classification accuracies. On some data sets it even achieves better accuracies. On ENZYMES the 3-LWL-APPROX is about 6% worse than the exact algorithm. But it still achieves better classification accuracies than the other kernels.

**A5** Both, the linear algebra based algorithm and the parallel variants, greatly improve the running time. For example on the PROTEINS data set the 3-LWL-LA (3-LWL-APPROX-PAR with inital sample size 100 and $\varepsilon = 0.05$) is more than five (three) times faster than the 3-LWL (3-LWL-APPROX).

VI. CONCLUSION AND FUTURE WORK

We proposed a graph kernel based on a local variant of the (global) $k$-dimensional Weisfeiler algorithm, which explores the space between local and global graph properties. We demonstrated that it can be computed approximately in constant time for bounded-degree graphs. For general graphs we proposed an adaptive sampling algorithm. Our experimental study showed that our kernel advances the state-of-the-art. In particular, the predictive accuracy of our kernel is favorable on all data sets which is not the case for the other kernels. We can draw the following conclusion:

> *The local $k$-LWL is able to extract* local as well as global *graph properties, and behaves favorably for all data sets. Moreover, it can be approximated efficiently. For bounded-degree graphs the approximation can be computed in* constant *time.*

Our work provides several interesting directions for future work. First, we believe that the space between local and global graph properties provides a growth path for designing novel graph kernels.

Secondly, it should be combined with the recent progress in deep learning approaches for graph classification or regression [56]–[59], as our sampling is a form of dropout. Moreover, the running time should be reduced further, e.g., by random walks [10], by further engineering, and more advanced techniques for bounding the conditional Rademacher average [50].

ACKNOWLEDGEMENT

This work has been supported by the German Science Foundation (DFG) within the Collaborative Research Center SFB 876 "Providing Information by Resource-Constrained Data Analysis", project A6 "Resource-efficient Graph Mining".

TABLE II: Classification accuracies in percent and standard deviations, OOT— Computation did not finish within one day, OOM— Out of memory.

|  | **Graph Kernel** | **Data Set** | | | | | | | |
|---|---|---|---|---|---|---|---|---|---|
|  |  | ENZYMES | IMDB-BINARY | MUTAG | NCI1 | NCI109 | PTC_FM | PROTEINS | REDDIT-BINARY |
| Local | GRAPHLET | 41.0 ±1.2 | 59.4 ±0.4 | **87.7** ±1.4 | 72.1 ±0.3 | 72.3 ±0.2 | 58.3 ±1.6 | 72.9 ±0.3 | 60.1 ±0.2 |
| Local | SHORTEST-PATH | 42.3 ±1.3 | 59.2 ±0.3 | 81.7 ±1.3 | 74.5 ±0.3 | 73.4 ±0.1 | 62.1 ±0.9 | **76.4** ±0.4 | **84.7** ±0.2 |
| Local | 1-WL | 53.4 ±1.4 | 72.4 ±0.5 | 78.3 ±1.9 | **83.1** ±0.2 | **85.2** ±0.2 | 62.9 ±1.6 | 73.7 ±0.5 | 75.3 ±0.3 |
| Glob. | 2-GWL | 49.7 ±1.6 | 71.5 ±0.8 | 83.6 ±1.6 | 71.3 ±0.3 | 70.8 ±0.4 | **64.7** ±0.2 | 75.2 ±0.4 | 67.0 ±0.1 |
| Glob. | 3-GWL | 49.6 ± 1.3 | 73.4 ±0.8 | 87.2 ±0.8 | 73.3 ±0.2 | 72.4 ±0.2 | 63.9 ±0.5 | OOT | OOM |
| Loc.+Glob. | 2-LWL | 51.6 ±0.8 | 72.1 ±0.4 | 85.2 ±1.6 | 77.0 ±0.2 | 76.9 ±0.2 | **64.7** ±0.2 | 75.0 ±0.3 | 75.0 ±0.4 |
| Loc.+Glob. | 3-LWL | **61.8** ±1.2 | **73.5** ±0.5 | 83.2 ±1.7 | **83.1** ±0.2 | 82.0 ±0.3 | 61.5 ±1.4 | 74.7 ±0.7 | OOM |
| Loc.+Glob. | 3-LWL-APPROX ($\varepsilon = 0.05$) | 55.0 ±0.9 | 71.4 ±1.0 | 80.2 ±1.7 | 78.3 ±0.4 | 77.0 ±0.3 | 64.2 ±0.8 | 76.0 ±0.6 | 81.3 ±0.5 |
| Loc.+Glob. | 3-LWL-APPROX ($\varepsilon = 0.1$) | 53.9 ±1.4 | 71.9 ±0.9 | 81.8 ±1.9 | 77.1 ±0.4 | 75.3 ±0.5 | 64.0 ±0.7 | **76.4** ±0.7 | 79.3 ±0.5 |

TABLE III: Average running times in seconds (Number of iterations for 1-WL, 3-LWL, 3-LWL-APPROX, and 3-GWL: 5, OOT— Computation did not finish within one day, OOM— Out of memory.

| **Graph Kernel** | **Data Set** | | | | | | | |
|---|---|---|---|---|---|---|---|---|
|  | ENZYMES | IMDB-BINARY | MUTAG | NCI1 | NCI109 | PTC_FM | PROTEINS | REDDIT-BINARY |
| GRAPHLET | <1 | <1 | <1 | <1 | <1 | <1 | <1 | 3.3 |
| SHORTEST-PATH | <1 | <1 | <1 | 2.3 | 2.1 | <1 | <1 | 1 115.6 |
| 1-WL | <1 | <1 | <1 | 1.6 | 1.6 | <1 | <1 | 3.1 |
| 2-LWL | <1 | 1.1 | <1 | 4.5 | 4.5 | <1 | 5.2 | 3 726.4 |
| 2-LWL-LA | <1 | <1 | <1 | 3.8 | 3.8 | <1 | 2.9 | 1 234.1 |
| 2-GWL | 6.5 | 3.2 | 2.0 | 37.0 | 37.0 | <1 | 368.0 | OOT |
| 2-GWL-LA | 1.7 | <1 | <1 | 11.5 | 13.3 | <1 | 62.8 | OOM |
| 3-LWL | 66.0 | 111.1 | 1.6 | 263.7 | 258.8 | 3.2 | 7 080.0 | OOM |
| 3-LWL-PAR | 36.2 | 26.7 | 1.0 | 128.9 | 130.9 | 2.4 | 2 218.4 | OOM |
| 3-LWL-LA | 16.5 | 18.2 | <1 | 69.7 | 70.5 | <1 | 1 318.4 | OOM |
| 3-LWL-APPROX ($\varepsilon = 0.05$) | 70.3 | 413.3 | 9.0 | 465.8 | 473.4 | 13.6 | 242.0 | 1 832.2 |
| 3-LWL-APPROX-PAR ($\varepsilon = 0.05$) | 23.6 | 97.2 | 3.3 | 162.4 | 165.9 | 5.9 | 76.0 | 810.8 |
| 3-LWL-APPROX ($\varepsilon = 0.1$) | 33.5 | 192.5 | 1.9 | 112.9 | 113.9 | 2.9 | 56.2 | 436.5 |
| 3-LWL-APPROX-PAR ($\varepsilon = 0.1$) | 12.5 | 46.3 | <1 | 45.8 | 46.5 | 1.6 | 20.0 | 216.5 |
| 3-GWL | 335.6 | 119.2 | 2.7 | 1 253.2 | 1 238.9 | 8.7 | OOT | OOT |
| 3-GWL-LA | 56.0 | 23.7 | <1 | 243.4 | 239.3 | <1 | OOM | OOM |